\documentclass{article} 
\usepackage{iclr2025_conference,times}

\usepackage{amsmath,amsfonts,bm}

\def\eqref#1{equation~\ref{#1}}

\def\1{\bm{1}}

\DeclareMathAlphabet{\mathsfit}{\encodingdefault}{\sfdefault}{m}{sl}
\SetMathAlphabet{\mathsfit}{bold}{\encodingdefault}{\sfdefault}{bx}{n}

\usepackage{hyperref}
\usepackage{url}
\usepackage[utf8]{inputenc} 
\usepackage[T1]{fontenc}    
\usepackage{booktabs}       
\usepackage{amsfonts}       
\usepackage{nicefrac}       
\usepackage{microtype}      
\usepackage{xcolor}         
\usepackage{graphicx}
\usepackage{subcaption}
\usepackage{multirow}
\usepackage{amsmath}
\usepackage[capitalize]{cleveref}
\definecolor{tabhighlight}{HTML}{e5e5e5}
\usepackage{color, colortbl}

\title{Tree of Attributes Prompt Learning for Vision-Language Models}

\author{
\begin{tabular}{@{}l}
Tong Ding$^{1,2}$\quad Wanhua Li$^1$\thanks{Corresponding Author}\quad Zhongqi Miao$^3$\quad Hanspeter Pfister$^1$ \\
\end{tabular}\\
$^1$Harvard University\quad $^2$Mass General Brigham\quad $^3$Microsoft
}
\newcommand{\ours}{TAP }

\iclrfinalcopy 
\begin{document}

\maketitle

\begin{abstract}
Prompt learning has proven effective in adapting vision language models for downstream tasks. However, existing methods usually append learnable prompt tokens solely with the category names to obtain textual features, which fails to fully leverage the rich context indicated in the category name. To address this issue, we propose the Tree of Attributes Prompt learning (TAP), which first instructs LLMs to generate a tree of attributes with a ``concept - attribute - description'' structure for each category, and then learn the hierarchy with vision and text prompt tokens. Unlike existing methods that merely augment category names with a set of unstructured descriptions, our approach essentially distills structured knowledge graphs associated with class names from LLMs. Furthermore, our approach introduces text and vision prompts designed to explicitly learn the corresponding visual attributes, effectively serving as domain experts. Additionally, the general and diverse descriptions generated based on the class names may be wrong or absent in the specific given images. To address this misalignment, we further introduce a vision-conditional pooling module to extract instance-specific text features. Extensive experimental results demonstrate that our approach outperforms state-of-the-art methods on the zero-shot base-to-novel generalization, cross-dataset transfer, as well as few-shot classification across 11 diverse datasets. Code is available at \href{https://github.com/HHenryD/TAP}{https://github.com/HHenryD/TAP}.
\end{abstract}

\section{Introduction}

Recent advancements in vision-language models (VLMs) like CLIP~\citep{radford2021learning} and ALIGN~\citep{jia2021scaling} merge the capabilities of visual perception with linguistic understanding, which have revolutionized the landscape with their zero-shot learning abilities. They proficiently handle tasks on unseen data, bypassing the conventional requirement for task-specific training. This feature has enabled a plethora of applications, ranging from content-based image retrieval to complex visual question answering, setting new benchmarks in the domain. A crucial development in this domain is the concept of prompt learning, which has significantly influenced both natural language processing (NLP) \citep{lester2021power, li2021prefixtuning, DBLP:journals/corr/abs-2110-07602} and vision-only models \citep{jia2022visual, wang2022dualprompt, wang2022learning, zhang2022neural}. This approach leverages learnable prompts to guide model understanding, tailoring responses to specific tasks or datasets.

\begin{figure}[t]
  \centering
  \includegraphics[width=\textwidth]{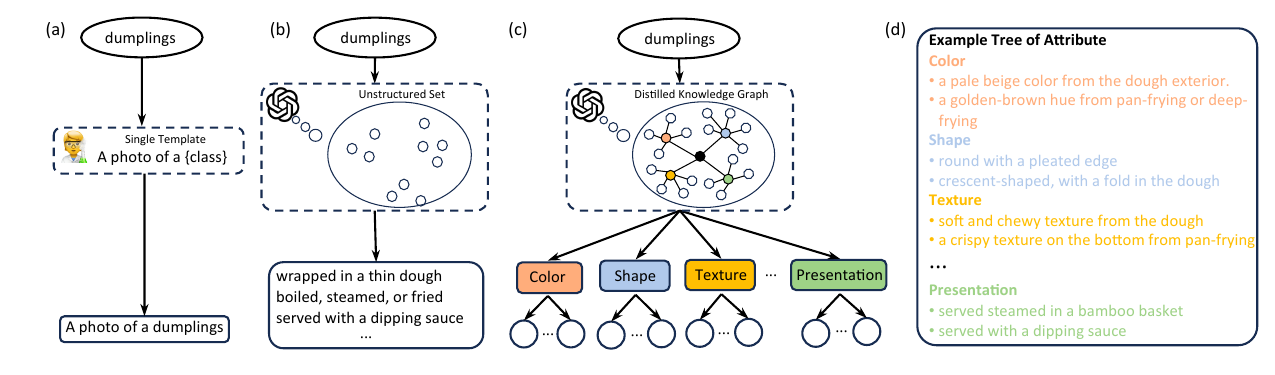}
  \caption{Illustration of the methods for CLIP text prompts formation. (a) Manually created prompt with the single ``a photo of a \{class\}'' template; (b) A unstructured set of detailed descriptions generated by LLMs; (c) The proposed Tree of Attribute distills a knowledge graph from LLMs, organizing the knowledge in ``concept - attribute - descriptions'' structure; (d) An example Tree of Attribute for class ``dumplings'', where each color represents a visual attribute.}
  \vspace{-15pt}
   \label{fig:overview}
\end{figure}

Prompt learning, particularly in vision-language models, has garnered considerable interest due to its parameter efficiency and rapid convergence \citep{zhou2022learning, zhou2022conditional, zhu2023prompt, derakhshani2023bayesian, lu2022prompt}. Techniques like CoOp \citep{zhou2022learning} optimize learnable continuous prompts for few-shot image recognition, enhancing model performance significantly. Recent efforts have expanded to multimodal prompt learning, optimizing prompts in both visual and language domains \citep{khattak2023maple, Khattak_2023_ICCV, shi2023logoprompt, lee2023read}. Despite their success, these models rely on simplistic text prompts, typically formatted as ``a photo of a \{class\}'', illustrated in \cref{fig:overview} (a). While functional, this approach lacks depth, failing to encapsulate the intricacies and finer details inherent in visual data. Such limitations hinder the model's ability to fully leverage the rich, descriptive potential offered by more detailed and contextually relevant textual information.

In parallel, another stream of research has been exploring the utilization of large language models (LLMs) to generate more elaborate and descriptive text prompts for enhancing zero-shot learning capabilities \citep{menon2022visual, pratt2023does, Roth_2023_ICCV,li2024SocialGPT, kim2023exposing, parkhi2012cats, yan2023learning, yang2023language, roy2024consistencyguided, zheng2023large, tian2023argue}. These LLM-generated descriptions offer a wealth of detail and context, potentially enriching the model's interpretative capabilities. However, current methodologies in integrating these descriptions often do not exploit the full potential of this richness. As shown in \cref{fig:overview} (b), most of these approaches lack a structured framework to organize and utilize these descriptions effectively, leading to a scattergun approach where not all generated descriptions are contextually relevant or optimally aligned with the visual content. In addition, as noted in~\citep{Roth_2023_ICCV}, descriptions generated by such paradigms are usually diverse, which covers most possibilities of the class, but include descriptions that are either likely not co-occurring, e.g. ``steamed'' and ``fried'', or absent in the input image, e.g. ``long tail'' for a cat shot from the front, necessitating the need for a selective pooling mechanism for clearer image-text alignments.

In response to these challenges, our work introduces ``Tree of Attribute Prompt learning (TAP),'' a method that redefines the integration and utilization of detailed descriptions within VLMs. As indicated in \cref{fig:overview} (c), unlike existing methods that merely augment category names with a set of unstructured descriptions, our approach essentially distills structured knowledge graphs associated with class names from LLMs. Specifically, we adopt a hierarchical, tree-like structure to systematically generate and integrate descriptions, ensuring a layered and comprehensive understanding of visual content. Each branch of this tree represents a specific attribute, with finer details fleshed out in the subsequent leaves, ensuring that every aspect of the visual content is captured and represented. Furthermore, we reimagine the learnable prompt tokens as ``domain experts'', each specializing in different aspects of the image, supplemented by the CLS token's global perspective. In addition, we introduce vision-conditional layers for each expert-attribute pair, which pool the most applicable descriptions from each of the attribute sets with condition on the input image content, ensuring optimal image-text alignment. This setup not only provides a detailed, attribute-focused analysis but also harmonizes these insights with the overall context.

Extensive experiments in base-to-novel generalization, cross-dataset transfer, and few-shot classification across 11 diverse datasets demonstrate the effectiveness of our method. On base-to-novel generalization, \ours achieves average performance gains of $1.07\%$ in harmonic mean over the state-of-the-art methods, and $9.34\%$ over the vanilla CLIP. On cross-dataset transfer, \ours outperforms existing methods on both source and target datasets by $1.03\%$ and $0.75\%$ in average. Competitive results are also observed in few-shot classification. 

\section{Related Work}
\textbf{Prompt Learning for Vision-Language Models.} Prompt learning bridges linguistic understanding and visual perception, originating in NLP~\citep{lester2021power, li2021prefixtuning, DBLP:journals/corr/abs-2110-07602} and later adapted to vision-only~\citep{jia2022visual, wang2022dualprompt, wang2022learning, zhang2022neural} and multimodal settings~\citep{zhou2022learning, zhou2022conditional, khattak2023maple, Khattak_2023_ICCV, shi2023logoprompt, lee2023read, tian2023argue, rasheed2023fine, roy2024consistencyguided, zheng2023large, zhu2023prompt, bulat2023lasp, lu2022prompt}. CoOp~\citep{zhou2022learning} introduced learnable continuous prompts for few-shot image recognition but struggled to generalize to unseen classes, highlighting the challenge of base-to-novel generalization~\citep{zhou2022conditional, guo2024scaffold, hernandez2024conformal, guo2024umap}. CoCoOp~\citep{zhou2022conditional} addressed this by conditioning prompts on visual features. Visual and multimodal prompt tuning methods, such as VPT~\citep{bahng2022visual}, DPT~\citep{10171397}, and MaPLe~\citep{khattak2023maple}, optimize prompts in pixel or text space to enhance feature alignment. Other works~\citep{Khattak_2023_ICCV, bulat2023lasp, li2022ordinalclip, roy2024consistencyguided} focus on regularization to improve generalization. PromptSRC introduced self-regulating prompts to better retain base knowledge, while methods like PLOT~\citep{chen2023plot} and ALIGN~\citep{wang2023tuning} apply Optimal Transport to align prompts with local features. Our work differs by introducing a hierarchical "Tree of Attribute" framework to structure textual descriptions and guide specialized "domain expert" tokens for attribute-level understanding.

\textbf{Enhancing model's understanding using visual attributes.} There’s a growing emphasis on the use of detailed visual descriptions for various visual understanding tasks, including more fine-grained captioning~\citep{GBC2024}, identifying subordinate-level categories~\citep{liu2024democratizing}, and language-guided visual classification~\citep{menon2022visual}. While manual creation is impractical given the large number of image classes, existing research relies either on data augmentation~\citep{kim2024aapl} or generation by LLMs such as GPT-3~\citep{brown2020language}, which offers an efficient generation of a broad spectrum of class-specific descriptions. These descriptions, like ``fur pattern'' or ``tail shape'' of a cat, provide fine-grained and distinctive characteristics. In an essence, such approaches can be viewed as knowledge distillation from LLMs trained on trained on vast and diverse textual corpora. However, existing studies often lack a structured methodology for distillation~\citep{kim2023exposing, menon2022visual, parkhi2012cats, Roth_2023_ICCV, yan2023learning, yang2023language, fabian2023multimodal, pratt2023does, novack2023chils, mao2023doubly, tian2023argue, zheng2023large, Zhang_Zhang_Yu_Tang_He_2024, liu2024multi} or fail to effectively exploit the inherent hierarchy within the knowledge~\citep{maniparambil2023enhancing, wang2024learning, GBC2024, liu2024democratizing}. Our approach (\ours) addresses these limitations by introducing a novel method to distill a knowledge graph from LLMs in a top-down manner, transitioning from class names (concepts) to visual attributes (e.g., color, shape) and further to detailed descriptions of each attribute, forming a structured Tree of Attributes (ToA). To fully leverage the ToA, we propose a bottom-up integration pipeline. We introduce vision-conditional pooling (VCP) layers to aggregate descriptions into attribute-level features, effectively mitigating potential noise in the generated descriptions. The alignment between attributes and introduced visual expert tokens is then refined through this hierarchical structure. This integration enables the model to exploit structured relationships within the ToA, enhancing both the granularity and interpretability of vision-text alignment.

\section{Methodology}

\subsection{Preliminary}

\textbf{CLIP.} Our approach is built on the pre-trained vision-language model, CLIP \citep{radford2021learning}. Formally,  let ${(x, c)}$ denote the dataset, where $x$ is an image and $c \in \{1, \ldots, C\}$ are the class labels. For an image $x$, the vision encoder $h_I(\cdot)$ transforms it into a feature vector $\bm{f}^v_x = h_I(x)$. Simultaneously, each class label $c$ is mapped to a text prompt $t_c=\texttt{a photo of a \{c\}}$, and converted into textual feature vectors ${\bm{f}^t_c} =h_T(t_c)$. The predicted class $\hat{y}$ is given by:
\begin{equation}
\hat{y} = \underset{c}{\mathrm{argmax}} \cos(\bm{f}^v_x, \bm{f}^t_c)
\end{equation}
where $\cos(\cdot)$ denotes cosine similarity.

\textbf{Image classification with class descriptions.} To improve the model's understanding of the categories in the transfer datasets, previous works~\citep{menon2022visual, Roth_2023_ICCV} use more detailed descriptions from Large Language Models (LLMs) instead of the simple \texttt{"a photo of a \{c\}"} to prompt the CLIP text encoder. Under this approach, a convoluted set of descriptions is generated for a class $c$ as $\mathbb{D}_c: \{$\texttt{"c, which is/has/etc description."} $\}$, e.g. \texttt{c}=\texttt{"television"} and \texttt{description}=\texttt{"black or grey"}. This classification is reformulated as
\begin{equation}
    \hat{y} = \underset{c}{\mathrm{argmax}} \frac{1}{|\mathbb{D}_c|} \underset{d\in \mathbb{D}_c}{\sum} \cos(h_I(x), h_T(d))
\end{equation}

\begin{figure}[t]
  \centering
  \includegraphics[width=\textwidth]{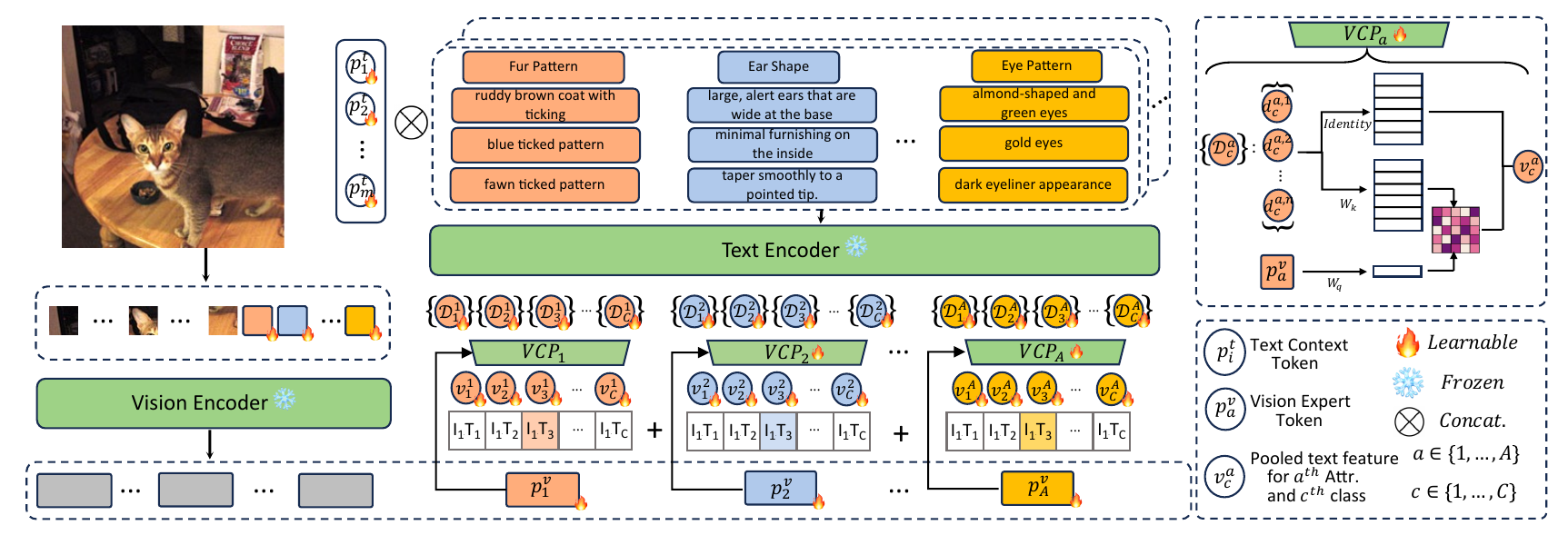}
  \caption{Overview of the proposed \ours method. \ours uses a bottom-up approach to aggregate the generated Tree of Attribute. The vision-conditional pooling (VCP) layer aggregates descriptions into attribute-level features, which are aligned with visual expert tokens focusing on specific attributes (e.g., color, texture). These attribute-level features are then combined to make class predictions via a weighted sum of logits from each attribute, fully leveraging the hierarchical structure within the tree.}
  \vspace{-15pt}
   \label{fig:arch}
\end{figure}

\subsection{Overall Framework}

We rethink the descriptions by LLM $\mathbb{D}_c$ as nodes in knowledge graphs. While previous methods generate an unstructured set of descriptions, we distill structured knowledge graphs for each class $c$ from LLM, in which the root node is the class name $c$, capturing the highest level semantics, and the leaf nodes are the detailed descriptions capturing fine-grained details. In this framework, previous paradigms only generate the leaf nodes of the graph, with the edges and graph structure missing, where the rich and inherent structure from the descriptions is overlooked. To address this limitation, we formulate our approach as a Tree of Attribute, which follows the ``concept - attribute - description'' structures, as illustrated in \cref{fig:overview} (c). 

Besides weighting the descriptions equally, previous works align descriptions that describe images from different aspects and at different granularities with a singular CLS token from the image encoder. However, while the use of a single CLS token is effective in certain contexts, we note that the CLS token is designed to capture the global information of an input image $x$~\citep{dosovitskiy2021an}. As a result, even though this helps to further inform global understanding, it may fail to effectively capture the nuances and variances at the attribute level, which leads to suboptimal use of the rich descriptions. We address this by introducing a set of learnable prompt tokens that serve as domain experts in the vision branch, each of which aligns with a specific attribute-level textual embedding.

Additionally, close inspection of the LLM-generated descriptions indicates limited contextual relevance and a high degree of diversity. Previous works~\citep{Roth_2023_ICCV} reflect the issue of descriptions that are likely not co-occurring e.g. ``steam'' and ``fried''. We further identify cases where the descriptions are technically correct but irrelevant to certain images, such as describing ``long tail'' in frontal images of cats, underscoring the need for a selective pooling mechanism. Thus, we introduce a vision-conditional pooling layer to extract instance-specific text features for each attribute for selecting the most applicable descriptions. 

Overall, \ours leverages the tree structure in two key ways: first, a top-down process generates attributes and corresponding descriptions for each class in a structured and contextually relevant manner. This ensures that the descriptions are structured and contextually relevant. Second, a bottom-up process aggregates information from the leaf nodes (descriptions) into attribute-level features, which are aligned with visual expert tokens. These expert tokens focus on fine-grained visual attributes, such as color or shape. Finally, the aggregated attribute-level features contribute to class predictions using a weighted sum of prediction logits, fully utilizing the hierarchical relationships within the tree. This dual approach allows \ours to capture both high-level structure and fine-grained details, leading to enhanced alignment of visual and textual data and improved model performance and interpretability. Inspired by CoOP \citep{zhou2022learning}, we also incorporate textual contextual tokens in the text encoder. The overall framework is presented in \cref{fig:arch}.

\subsection{Tree of Attribute generation by LLMs} 
We redefine the process of integrating LLM-generated descriptions by introducing a knowledge graph $\mathcal{G}_c=\{\mathbb{V}_c, \mathbb{E}_c\}$ for each class $c$, where $\mathbb{V}_c$ denotes the set of nodes, and $\mathbb{E}_c$ denotes the edges that capture the semantic relationship between nodes. In previous works, $\mathbb{V}_c$ is the set of descriptions $\mathbb{D}_c$, while $\mathbb{E}_c$ is missing. We argue that such methods overlook the inherent structure among the descriptions and thus do not exploit the richness of these descriptions effectively. To better leverage knowledge from LLMs, we introduce an attribute layer to link the root node class name, and the leaf node descriptions. The attribute nodes include visual attributes generated by LLMs, such as color and shape, for systematically guiding description generation as illustrated in ~\cref{fig:overview} (c). Each branch of this ``tree'' represents a specific attribute, with the subsequent ``leaves'' fleshing out the descriptions with finer details. In this framework, $\mathbb{G}_c$ includes the class name which is the root node, the set of attributes such as color and shape being the intermediate layer, and lastly the set of descriptions under each attribute node. $\mathbb{E}_c$ includes the edges that build up the hierarchy. This structure allows for a nuanced representation of class information, spanning from general concepts down to specific attributes and detailed descriptions.

To this end, we introduce the Tree of Attribute (ToA), where we use a tree structure to model the relationship and structure of the descriptions. Let $\mathbb{A}_c$ denote the set of attributes, and for each attribute $a_c\in \mathbb{A}_c$, we denote its leaf nodes as $\mathbb{D}^a_c$. Each set $\mathbb{D}_c^a$ contains descriptions that specifically pertain to attribute $a$ for class $c$, which is denoted as 
\begin{equation}
\mathbb{D}_c^a = \{d_c^{a,1}, d_c^{a,2}, \ldots, d_c^{a,n}\},
\end{equation}
where $d_c^{a,i}$ represents the $i$-th description for attribute $a$ of class $c$ and $n$ is the number of descriptions per attribute.

The process of generating a Tree of Attribute (ToA) unfolds in three steps:
1) \textbf{Attribute Generation:} We first query LLMs with the dataset information and ask it to generate a set of attributes $\mathbb{A}$ which are considered relevant and characteristic of the dataset. 2) \textbf{Example Generation:} We then ask LLMs to generate descriptions for a randomly sampled class in the dataset, using the attributes $\mathbb{A}$ identified in the previous step. Each description takes the format of ``class, which \{is/has/etc\} \{description\}''. 3) \textbf{Description Generation for All Classes:} Building upon the Q\&A template from the previous step, the LLM is then tasked with generating descriptions for all classes in the dataset. 

Additionally, we incorporate a ``global context'' attribute which is aligned with the CLS token in the vision encoder. The descriptions are the 7 standard templates provided in \citep{radford2021learning}.

\subsection{Learning \ours with Learnable Expert Tokens} 
To fully exploit the structured Tree of Attribute, we introduce learnable visual expert tokens $\textbf{p}_a^v$ in the vision branch to learn from each of the attribute nodes $a\in \mathbb{A}$. Unlike traditional methods that rely on a single CLS token for alignment, these expert tokens enable focused learning on specific image attributes, such as color or shape, enhancing the model's performance and interpretability.

We denote the set of introduced visual expert tokens as $\mathbb{P}^v=\{\bm{p}^v_a|a\in \mathbb{A}\}$. Akin to the idea of visual prompt tuning (VPT)~\citep{jia2022visual}, we insert $\mathbb{P}^v$ into the input sequence of the vision encoder, forming the prompted input sequences $\mathbb{\Tilde{X}}_p=\{\bm{e}_{\text{CLS}}, \mathbb{P}^v,\mathbb{E}_{\text{patch}}\}$, where $\bm{e}_{\text{CLS}}$ is the input CLS token, and $\mathbb{E}_{\text{patch}}$ denotes the embedded patch tokens. To further boost the model's capacity for nuanced attribute representation, we employ deep prompting by introducing a zero-initialized layer residual for each prompt token across transformer layers, which provides more explicit attribute guidance across transformer layers. In parallel, we adopt a set of $m$ learnable context tokens $\mathbb{P}^t=\{\bm{p}^t_j|j\in \{1,2,...,m\}\}$ for the text encoder shared across all descriptions, similar to \citep{zhou2022learning}. 

\subsection{Vision-Conditional Pooling} 
To mitigate issues of misalignment and potential misleading information from the broad spectrum of LLM-generated descriptions, we proposed an adaptive vision-conditional pooling layer, applicable to each set of attribute descriptions $\mathbb{D}_a$ shared across all classes to dynamically pool the most applicable descriptions based on the visual content of the image $x$ using its corresponding visual expert token denoted as $\bm{p}_{a,x}^v$. For ease of expression, we will proceed without explicitly mentioning $x$, though it's important to note that both the expert token and the resulting attribute-level embeddings are dependent on the visual information. Intuitively, VCP uses attention to calculate the similarity between $\bm{p}_{a}^v$ and all embedded descriptions in attribute $\mathbb{D}_a$, which are then used as weights for a weighted sum of the original description embeddings. Formally, for each attribute $a$ and its associated expert token $\bm{p}_{a}^v$, the pooled attribute-level embedding $\bm{v}^a_c$ for class $c$ and attribute $a$ is:
\begin{equation}
    \begin{aligned}
    \text{Query} &= \bm{W_q} \cdot \bm{p}_a^v, \\
    \text{Key} &= \bm{W_k} \cdot \texttt{Emb}(\mathbb{D}^a_c), \\
    \text{Attention Score} &= \texttt{softmax}(\text{Query} \cdot \text{Key}^T), \\
    \bm{v}^a_c &= \bm{\text{Attention Score}} \cdot \texttt{Emb}(\mathbb{D}^a_c),
    \end{aligned}
    \label{eq:vcp}
\end{equation}
where $W_q$ and $W_k$ are learnable weights $\in \mathbb{R}^{d \times d}$, $\texttt{Emb}(\cdot)$ denotes the embedding function, and $\texttt{softmax}(\cdot)$ is the Softmax function. This layer mirrors cross-attention but omits $\bm{W_v}$ to maintain the output within the CLIP V-L space.

\subsection{Training and Inference}

\textbf{Training objective.} 
During training, each visual expert token $\textbf{p}_a^v$ is aligned with its associated attribute-level embedding $\textbf{v}^a_c$, trained with the following contrastive objective:
\begin{equation}
L_{{con}}(\bm{p}_a^v, \bm{v}^a_c) = -\frac{1}{N} \sum_{i=1}^N \log \frac{\exp(\cos(\bm{p}_a^v, \bm{v}_y^a)/\tau)}{\sum_{c=1}^C \exp(\cos(\bm{p}_{a}^v, \bm{v}_c^{a})/\tau)},
\end{equation}
where $N$ represents the number of training samples, and $\tau$ is the learned temprature of CLIP. The total classification loss $L_{\text{class}}$ is the average of the contrastive loss from each expert token as well as the CLS token, defined as: 
\begin{equation}
L_{{class}} = \frac{1}{|\mathbb{A}|} \bigg (\sum_{a\in \mathbb{A}} L_{{con}}(\bm{p}_a^v, \bm{v}_c^a))\bigg ),
\end{equation}

Similar to \citep{Khattak_2023_ICCV} and \citep{bulat2023lasp}, we regularize the vision CLS token, text feature, and the prediction logits from each attribute using the vanilla CLIP model. We denote the regularization loss as $L_{reg}$, where the details can be found in Appendix. The overall training objective is $L_{\text{total}}=L_{\text{class}}+L_{\text{reg}}$.

\textbf{Prediction fusion.} During inference, we integrate the prediction by each attribute expert pair by a weighted sum, formulated as follows:
\begin{equation}
\label{eq:fusion}
\Tilde{y} = \underset{c}{\mathrm{argmax}} \bigg( \alpha \cos(\bm{f}^v_{{CLS}}, \bm{v}_{c}^{CLS})
+ \frac{1-\alpha}{|\mathbb{A}|-1} \sum_{a \in \mathbb{A}\textbackslash \{CLS\}} \cos(\bm{p}_{a}^v, \bm{v}_{c}^{a}) \bigg)
\end{equation}
where $\alpha$ is a hyperparameter that signifies the weight assigned to the global context provided by the CLS token, balancing its contribution with that of the attribute-specific expert prompts.

\begin{table}[!t] 
    \setlength{\tabcolsep}{1.5pt}
    \caption{Comparison with state-of-the-art methods in base-to-novel generalization. The model is trained on the base class, and evaluated on the unseen novel classes in zero-shot. \ours demonstrates strong generalization performance.  HM: harmonic mean~\citep{xian2017zero}. }
    \vspace{-5pt}
    \label{table:comparision_main}
    \begin{subtable}[t]{0.24\textwidth}
    \centering
    \caption{\textbf{Average}}
    \resizebox{0.99\linewidth}{!}{
    \label{table:average_acc}
    \begin{tabular}{l cc|c}
    \toprule
    & Base & Novel & HM \\
    \midrule
    CLIP & 69.34 & 74.22 & 71.70 \\
    CoOp &  {82.69} & 63.22 & 71.66 \\
    Co-CoOp & 80.47 & 71.69 & 75.83 \\
    ProGrad & 82.48 & 70.75 & 76.16 \\
    RPO & 81.13 & 75.00 & 77.78 \\
    LoGoPrompt & 84.47 & 74.24 & 79.03 \\
    PromptSRC & 84.26 & 76.10 & 79.97 \\
    \rowcolor{tabhighlight}
    \ours & \textbf{84.75} & \textbf{77.63} & \textbf{81.04} \\
    \bottomrule
    \end{tabular}
    }
    \end{subtable}
    \begin{subtable}[t]{.24\textwidth}
    \centering
    \caption{ImageNet}
    \resizebox{0.99\linewidth}{!}{
    \begin{tabular}{l cc|c}
    \toprule
    & Base & Novel & HM \\
    \midrule
    CLIP & 72.43 & 68.14 & 70.22 \\
    CoOp & {76.47} & 67.88 & 71.92\\
    Co-CoOp & 75.98 & {70.43} & {73.10} \\
    ProGrad & 77.02& 66.66 &71.46\\
    RPO & 76.60 & 71.57 & 74.00 \\
    LoGoPrompt & 76.74 & \textbf{70.83} & 73.66 \\
    PromptSRC & 77.60 & 70.73 & \textbf{74.01}\\
    \rowcolor{tabhighlight}
    \ours & \textbf{77.97} & 70.40 & 73.99 \\ 
    \bottomrule
    \end{tabular}
    }
    \end{subtable}
    \begin{subtable}[t]{.24\textwidth}
    \centering
    \caption{Caltech101}
    \resizebox{0.99\linewidth}{!}{
    \begin{tabular}{l cc|c}
    \toprule
    & Base & Novel & HM \\
    \midrule
    CLIP & 96.84 & {94.00} & 95.40 \\
    CoOp & {98.00} & 89.81 & 93.73 \\
    Co-CoOp & 97.96 & 93.81 & {95.84} \\
    ProGrad & 98.02  & 93.89  & 95.91 \\ 
    RPO  & 97.97 & 94.37 & 96.03 \\ 
    LoGoPrompt & 98.19 & 93.78 & 95.93 \\
    PromptSRC & 98.10 & 94.03 & 96.02 \\ 
    \rowcolor{tabhighlight}
    \ours & \textbf{98.90} & \textbf{95.50} & \textbf{97.17} \\
    \bottomrule
    \end{tabular}
    }
    \end{subtable}
    \begin{subtable}[t]{.24\textwidth}
    \centering
    \caption{OxfordPets}
    \resizebox{0.99\linewidth}{!}{     
    \begin{tabular}{l cc|c}
    \toprule
    & Base & Novel & HM \\
    \midrule
    CLIP & 91.17 & 97.26 & 94.12 \\
    CoOp & 93.67 & 95.29 & 94.47 \\
    Co-CoOp & {95.20} & {97.69} & {96.43} \\
    ProGrad & 95.07 & 97.63&  96.33\\
    RPO & 94.63 & 97.50 & 96.05 \\ 
    LoGoPrompt & \textbf{96.07} & 96.31 & 96.18 \\
    PromptSRC & 95.33 & 97.30 & 96.30 \\ 
    \rowcolor{tabhighlight}
    \ours & 95.80 & \textbf{97.73} & \textbf{96.76} \\ 
    \bottomrule
    \end{tabular}
    }
    \end{subtable}
    \\
    \begin{subtable}[t]{.24\textwidth}
    \centering
    \caption{StanfordCars}
    \resizebox{0.99\linewidth}{!}{     
    \begin{tabular}{l cc|c}
    \toprule
    & Base & Novel & HM \\
    \midrule
    CLIP & 63.37 &  {74.89} & 68.65 \\
    CoOp & {78.12} & 60.40 & 68.13 \\
    Co-CoOp & 70.49 & 73.59 & {72.01} \\ 
    ProGrad  & 77.68  & 68.63  & 72.88 \\ 
    RPO  & 73.87 & \textbf{75.53} & 74.69 \\ 
    LoGoPrompt & 78.36 & 72.39 & 75.26 \\
    PromptSRC & 78.27 & 74.97 & 76.58 \\ 
    \rowcolor{tabhighlight}
    \ours & \textbf{80.70} & {74.27} & \textbf{77.35} \\ 
    \bottomrule
    \end{tabular}
    }
    \end{subtable}
    \begin{subtable}[t]{.24\textwidth}
    \centering
    \caption{Flowers102}
    \resizebox{0.99\linewidth}{!}{     
    \begin{tabular}{l cc|c}
    \toprule
    & Base & Novel & HM \\
    \midrule
    CLIP & 72.08 & \textbf{77.80} & 74.83 \\
    CoOp & {97.60} & 59.67 & 74.06 \\
    Co-CoOp & 94.87 & 71.75 & {81.71} \\ 
    ProGrad  & 95.54  & 71.87  & 82.03 \\ 
    RPO  & 94.13 & 76.67 & 84.50 \\ 
    LoGoPrompt & \textbf{99.05} & 76.52 & \textbf{86.34} \\
    PromptSRC & 98.07 & 76.50 & 85.95 \\ 
    \rowcolor{tabhighlight}
    \ours & {97.90} & {75.57} & {85.30} \\ 
    \bottomrule
    \end{tabular}
    }
    \end{subtable}
    \begin{subtable}[t]{.24\textwidth}
    \centering
    \caption{Food101}
    \resizebox{0.99\linewidth}{!}{     
    \begin{tabular}{l cc|c}
    \toprule
    & Base & Novel & HM \\
    \midrule
    CLIP & 90.10 & 91.22 & 90.66 \\
    CoOp & 88.33 & 82.26 & 85.19 \\
    Co-CoOp & {90.70} & {91.29} & {90.99} \\ 
    ProGrad & 90.37&  89.59 & 89.98 \\
    RPO & 90.33 & 90.83 & 90.58 \\
    LoGoPrompt & 90.82 & 91.41 & 91.11 \\
    PromptSRC & 90.67 & 91.53 & 91.10 \\ 
    \rowcolor{tabhighlight}
    \ours & \textbf{90.97} & \textbf{91.83} & \textbf{91.40} \\
    \bottomrule
    \end{tabular}
    }
    \end{subtable}
    \begin{subtable}[t]{.24\textwidth}
    \centering
    \caption{FGVCAircraft}
    \resizebox{0.99\linewidth}{!}{     
    \begin{tabular}{l cc|c}
    \toprule
    & Base & Novel & HM \\
    \midrule
    CLIP & 27.19 &  {36.29} & {31.09} \\
    CoOp & {40.44} & 22.30 & 28.75 \\
    Co-CoOp & 33.41 & 23.71 & 27.74 \\ 
    ProGrad & 40.54 & 27.57 & 32.82 \\
    RPO & 37.33 & 34.20 & 35.70 \\
    LoGoPrompt & \textbf{45.98} & 34.67 & 39.53 \\
    PromptSRC & 42.73 & \textbf{37.87} & \textbf{40.15} \\ 
    \rowcolor{tabhighlight}
    \ours & {44.40} & 36.50 & {40.06} \\ 
    \bottomrule
    \end{tabular}
    }
    \end{subtable}
    \\
    \begin{subtable}[t]{.24\textwidth}
    \centering
    \caption{SUN397}
    \resizebox{0.99\linewidth}{!}{     
    \begin{tabular}{l cc|c}
    \toprule
    & Base & Novel & HM \\
    \midrule
    CLIP & 69.36 & 75.35 & 72.23 \\
    CoOp & {80.60} & 65.89 & 72.51 \\
    Co-CoOp & 79.74 & {76.86} & {78.27} \\ 
    ProGrad & 81.26 & 74.17&  77.55 \\
    RPO & 80.60 & 77.80 & 79.18 \\
    LoGoPrompt & 81.20 & 78.12 & 79.63 \\
    PromptSRC & 82.67 & 78.47 & 80.52 \\ 
    \rowcolor{tabhighlight}
    \ours & \textbf{82.87} & \textbf{79.53} & \textbf{81.17} \\ 
    \bottomrule
    \end{tabular}
    }
    \end{subtable}
    \begin{subtable}[t]{.24\textwidth}
    \centering
    \caption{DTD}
    \resizebox{0.99\linewidth}{!}{     
    \begin{tabular}{l cc|c}
    \toprule
    & Base & Novel & HM \\
    \midrule
    CLIP & 53.24 & {59.90} & 56.37 \\
    CoOp & {79.44} & 41.18 & 54.24 \\
    Co-CoOp & 77.01 & 56.00 & {64.85} \\ 
    ProGrad & 77.35&  52.35&  62.45 \\
    RPO & 76.70 & 62.13 & 68.61 \\ 
    LoGoPrompt & 82.87 & 60.14 & 69.70 \\
    PromptSRC & 83.37 & 62.97 & 71.75 \\ 
    \rowcolor{tabhighlight}
    \ours & \textbf{84.20} & \textbf{68.00} & \textbf{75.24} \\ 
    \bottomrule
    \end{tabular}
    }
    \end{subtable}
    \begin{subtable}[t]{.24\textwidth}
    \centering
    \caption{EuroSAT}
    \resizebox{0.99\linewidth}{!}{     
    \begin{tabular}{l cc|c}
    \toprule
    & Base & Novel & HM \\
    \midrule
    CLIP & 56.48 & {64.05} & 60.03 \\
    CoOp & {92.19} & 54.74 & 68.69 \\
    Co-CoOp & 87.49 & 60.04 & {71.21} \\ 
    ProGrad & 90.11 & 60.89 & 72.67 \\ 
    RPO & 86.63 & 68.97 & 76.79 \\ 
    LoGoPrompt & \textbf{93.67} & 69.44 & 79.75 \\
    PromptSRC & 92.90 & 73.90 & 82.32 \\ 
    \rowcolor{tabhighlight}
    \ours & 90.70 & \textbf{82.17} & \textbf{86.22} \\ 
    \bottomrule
    \end{tabular}
    }
    \end{subtable}
    \begin{subtable}[t]{.24\textwidth}
    \centering
    \caption{UCF101}
    \resizebox{0.99\linewidth}{!}{     
    \begin{tabular}{l cc|c}
    \toprule
    & Base & Novel & HM \\
    \midrule
    CLIP & 70.53 & {77.50} & 73.85 \\
    CoOp & {84.69} & 56.05 & 67.46 \\
    Co-CoOp & 82.33 & 73.45 & {77.64} \\ 
    ProGrad & 84.33 & 74.94 & 79.35 \\ 
    RPO & 83.67 & 75.43 & 79.34 \\ 
    LoGoPrompt & 86.19 & 73.07 & 79.09 \\
    PromptSRC & 87.10 & 78.80 & 82.74 \\ 
    \rowcolor{tabhighlight}
    \ours & \textbf{87.90} & \textbf{82.43} & \textbf{85.08} \\ 
    \bottomrule
    \end{tabular}
    }
    \end{subtable}
    \vspace{-15pt}
\end{table}

\section{Experiments}

We extensively evaluate our method in three settings: 1) Base-to-novel class generalization, where the datasets are equally split into base and novel classes. We train the model on the base classes only and evaluate on both base and novel classes; 2) Cross-dataset transfer, where we train on ImageNet with 16 shots per class, and directly evaluate on other datasets in zero-shot; and 3) Few-shot classification with 16 shots per class. 

\textbf{Datasets and baslines.} For all of the three settings, we follow previous works~\citep{zhou2022learning, zhou2022conditional}, using 11 image recognition datasets, including: ImageNet~\citep{deng2009imagenet} and Caltech101~\citep{fei2004learning} for generic object recognition; OxfordPets~\citep{parkhi2012cats}, StanfordCars~\citep{krause20133d}, Flowers102~\citep{nilsback2008automated}, Food101~\citep{bossard2014food}, and FGVCAircraft~\citep{maji2013fine} for fine-grained classification; SUN397~\citep{xiao2010sun} for scene recognition; UCF101~\citep{soomro2012ucf101} for action recognition; DTD~\citep{cimpoi2014describing} for
texture classification; and EuroSAT~\citep{helber2019eurosat} for satellite image analysis. We benchmark against several leading methods, including CLIP \citep{radford2021learning}, CoOp \citep{zhou2022learning}, Co-CoOP \citep{zhou2022conditional}, ProGrad \citep{zhu2023prompt}, RPO \citep{lee2023read}, LoGoPrompt \citep{shi2023logoprompt}, and the state-of-the-art PromptSRC \citep{Khattak_2023_ICCV}.

\textbf{Implementation details.}
A pre-trained CLIP model with a ViT-B/16 vision backbone is used in all of our experiments and results are averaged over 3 runs. We use GPT-3.5-turbo~\citep{ouyang2022training} for attribute and description generation. We initialize the text context tokens with the word embedding of \texttt{"a photo of a."} During training, we iteratively train the vision and text encoders with 5 epochs for vision and 1 epoch for text schedule. We train a total of 60, 24, and 120 epochs for base-to-novel generalization, cross-dataset transfer, and few-shot classification respectively. We set $\alpha=0.4$ for all datasets. We also use a Gaussian Prompt Weighting (GPA) following \citep{Khattak_2023_ICCV}, with a mean of $0.9 N$, std of $0.1N$, where $N$ represents the total number of epochs, for all tasks. Refer to the Appendix for additional implementation details.

\subsection{Base-to-Novel Generalization}
In base-to-novel generalization, we equally split the classes into base and novel classes. Initial training and evaluations are conducted on the seen base classes, followed by evaluation on the unseen novel classes in a zero-shot manner. \ours surpasses prior state-of-the-art models in terms of the base and novel class accuracy, as well as their harmonic mean across most of the 11 datasets, with an average increase of 1.53\% in the zero-shot novel class prediction, and a 1.07\% increase in the overall harmonic mean in average, as detailed in \cref{table:comparision_main}. Notably, our method improves unseen class prediction without compromising base class performance, exhibiting an average performance boost of 0.49\%. In the challenging fine-grained tasks such as DTD, EuroSAT, and UCF101, \ours achieves significant improvements in novel class prediction by 5.03\%, 8.27\%, and 3.63\% respectively. These results underscore the robust generalizability and efficacy of our method across diverse scenarios.

\subsection{Cross-dataset Transfer} 

To further investigate the generalization capability of \ours, we train on ImageNet with 16 shots per class, and directly test on the other 10 datasets under zero-shot without further tuning. 
As shown in \cref{tab:cross-dataset}, \ours demonstrates better generalizability on 8/10 target datasets compared to PromptSRC \citep{Khattak_2023_ICCV}, and achieves an average performance increase of $0.75\%$. Additionally, while the performance increase of previous methods on target datasets come with costs on the source dataset ($-0.49\%$ for CoCoOP and $-0.24\%$ for PromptSRC) as compared to CoOP \citep{zhou2022learning}, \ours also outperform previous methods on the source dataset with $1.03\%$ increase compared to PromptSRC ($0.79\%$ incrase compared to CoOP), demonstrating \ours's robustness in domain generalization without sacrifice on  source dataset performance.

\begin{table}[!t]
    \caption{Comparison with state-of-the-art methods in cross-dataset transfer evaluation. The model is trained on the source dataset and evaluated on the target datasets in zero-shot.}
    \vspace{-5pt}
    \resizebox{\textwidth}{!}{
        \begin{tabular}{l c ccccccccccc}
        \toprule
        & \textbf{Source} & \multicolumn{11}{c}{\textbf{Target}} \\ \cmidrule(lr){2-2} \cmidrule(lr){3-13} \\
        &  \footnotesize\rotatebox{60}{ImageNet} & \footnotesize\rotatebox{60}{Caltech101} & \footnotesize\rotatebox{60}{Pets} & \footnotesize\rotatebox{60}{Cars} & \footnotesize\rotatebox{60}{Flowers} & \footnotesize\rotatebox{60}{Food101} & \footnotesize\rotatebox{60}{Aircraft} & \footnotesize\rotatebox{60}{SUN397 } & \footnotesize\rotatebox{60}{DTD } & \footnotesize\rotatebox{60}{EuroSAT} & \footnotesize\rotatebox{60}{UCF101}  & 
        \footnotesize\rotatebox{60}{Average}\\
        \midrule
        CoOp & 71.51 & 93.70 & 89.14 & 64.51 & 68.71 & 85.30 & 18.47 & 64.15 & 41.92 & \textbf{46.39} & 66.55 & 63.88 \\
        CoCoOp & 71.02 & \textbf{94.43} & 90.14 & 65.32 & \textbf{71.88} & 86.06 & 22.94 & 67.36 & 45.73 & 45.37 & 68.21 & 65.74 \\
        PSRC & 71.27 & 93.60 & 90.25 & \textbf{65.70} & 70.25 & \textbf{86.15} & 23.90 & 67.10 & 46.87 & 45.50 & 68.75 & 65.81  \\
        \midrule
        \rowcolor{tabhighlight}
        \ours  & \textbf{72.30} & 94.30 & \textbf{90.70} & 65.60 & 70.93 & 86.10 & \textbf{24.57} & \textbf{68.30} & \textbf{50.20} & 46.00 & \textbf{68.90} & \textbf{66.56} \\
        \bottomrule
        \end{tabular}
        }
        \label{tab:cross-dataset} 
        \vspace{-5pt}
\end{table}

\begin{table}[!t]
    \caption{Comparison with state-of-the-art methods in few shot classification results with 16 shots.}
    \vspace{-5pt}
    \resizebox{\textwidth}{!}{
        \begin{tabular}{l c ccccccccccc}
        \toprule
        & \multicolumn{12}{c}{16-Shot Classification} \\ \cmidrule(lr){2-13}
        & \footnotesize\rotatebox{60}{Average} & \footnotesize\rotatebox{60}{ImageNet} & \footnotesize\rotatebox{60}{Caltech101} & \footnotesize\rotatebox{60}{Pets} & \footnotesize\rotatebox{60}{Cars} & \footnotesize\rotatebox{60}{Flowers} & \footnotesize\rotatebox{60}{Food101} & \footnotesize\rotatebox{60}{Aircraft} & \footnotesize\rotatebox{60}{SUN397 } & \footnotesize\rotatebox{60}{DTD } & \footnotesize\rotatebox{60}{EuroSAT} & \footnotesize\rotatebox{60}{UCF101}  \\
        \midrule
        CLIP  & 78.79  & 67.31 & 95.43 & 85.34 & 80.44 & 97.37 & 82.90 & 45.36 & 73.28 & 69.96 & 87.21 & 82.11 \\
        CoOp & 79.89 & 71.87 & 95.57 & 91.87 & 83.07 & 97.07 & 84.20 & 43.40 & 74.67 & 69.87 & 84.93 & 82.23 \\
        CoCoOp & 74.90 & 70.83 & 95.16 & 93.34 & 71.57 & 87.84 & 87.25 & 31.21 & 72.15 & 63.04 & 73.32 & 78.14 \\
        PSRC & 82.87 & 73.17 & 96.07 & 93.67 & 83.83 & 97.60 & 87.50 & \textbf{50.83} & 77.23 & 72.73 & \textbf{92.43} & 86.47  \\
        \midrule
        \rowcolor{tabhighlight}
        \ours  & \textbf{83.37} & \textbf{73.76} & \textbf{96.73} & \textbf{93.90} & \textbf{85.37} & \textbf{98.10} & \textbf{87.53} & 50.43 & \textbf{77.30} & \textbf{74.90} & 91.90 & \textbf{87.17} \\
    
        \bottomrule
        \end{tabular}
        }
        \label{tab:few-shot} 
        \vspace{-15pt}
\end{table}
\subsection{Few-Shot Classification} 

In few-shot classification, \ours also outperforms existing methods in 9 out of the 11 datasets.  Detailed in \cref{tab:few-shot}, we achieve an average accuracy of $83.37$ across the 11 datasets, surpassing the previous state-of-the-art methods by $0.5\%$, further demonstrating the effectiveness of our method.

\subsection{Ablation Study}

\begin{figure}[t]
\begin{minipage}[c]{0.65\textwidth}
    \includegraphics[width=\textwidth]{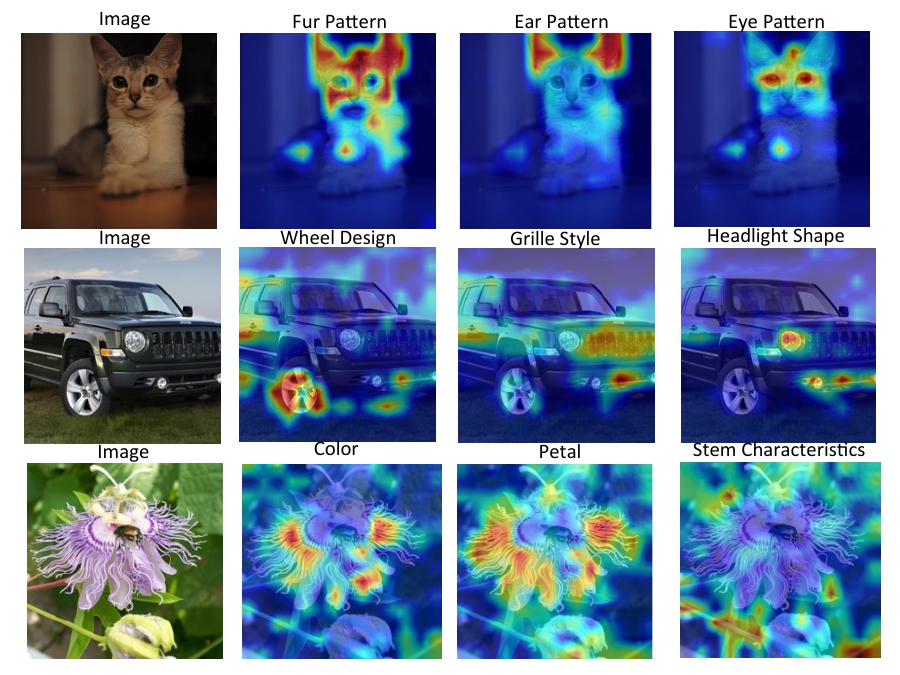}
    \caption{Visualization of the class activation maps.}
    \label{fig:vis_interpretablity}
\end{minipage}\hfill
\begin{minipage}[c]{0.35\textwidth}
    \centering
    \captionof{table}{Effects of the Tree of Attributes.}
    \label{table:toa}
    \begin{small}
    \begin{tabular}{c|cc}
        \toprule
        Des. Org. & Unstructured & Ours \\
        \midrule
        Base & 82.89 & \textbf{84.75} \\
        Novel & 75.32 & \textbf{77.63} \\
        HM & 78.93 & \textbf{81.04} \\
        \bottomrule
    \end{tabular}
    \end{small}
    \vspace{5pt} 
    \captionof{table}{Effects of domain experts.}
    \label{table:vision-expert}
    \begin{small}
    \begin{tabular}{c|cc}
        \toprule
        Align. Token & CLS & Ours \\
        \midrule
        Base & 83.89 & \textbf{84.75} \\
        Novel & 76.85 & \textbf{77.63} \\
        HM & 80.22 & \textbf{81.04} \\
        \bottomrule
    \end{tabular}
    \end{small}
\end{minipage}
\vspace{-15pt}
\end{figure}

\textbf{Effects of Tree of Attribute.} A core inquiry is whether structuring descriptions into a Tree of Attribute (ToA) offers advantages over an unstructured aggregation of LLM-generated descriptions. To evaluate, we revert to aligning a mixed, unstructured set of descriptions with the CLS token - a common practice in prior studies \citep{mao2023doubly, maniparambil2023enhancing, liu2024multi, wang2024learning, tian2023argue, zheng2023large}, while keeping the same number of visual prompt tokens. According to \cref{table:toa}, substituting the ToA with an unstructured set results in significant performance decreases of 1.86\%, 2.31\%, and 2.11\% across the average base, novel, and their harmonic mean performances, respectively. This stark contrast underscores the ToA's critical role in enhancing model efficacy.

\noindent \textbf{Effects of Learning through Domain Experts.} Further, we examine the impact of substituting the CLS token with visual expert tokens for learning fine-grained attributes, commonly adopted in in previous works \citep{mao2023doubly, lee2023read, tian2023argue, zheng2023large}. Our findings (\cref{table:vision-expert}) reveal improvements of 0.89\%, 0.78\%, and 0.82\% in the average base, novel, and harmonic mean accuracies, respectively, upon integrating visual expert tokens. These results support the notion that domain-specific, learnable tokens enhance the model's ability to grasp fine-grained details by focusing on distinct aspects of the image, as opposed to the CLS token's global focus.

\noindent \textbf{Effects of fusion coefficient $\alpha$.} $\alpha$ in \cref{eq:fusion} balance global and local information. We compare the performance of using CLS token only (i.e. $\alpha=1.0$) for making the final prediction against our proposed prediction fusion with $\alpha=0.4$. As shown in \cref{table:alpha}, using CLS token decreases the performance significantly on both base and novel classes. This result further demonstrates the limitations of using a singular CLS token which focuses on global information, and supports the effectiveness of the use of expert tokens which focus on local information.

\noindent \textbf{Effects of Number of Attributes.}
In our framework, the selection of attributes is dynamically determined by LLMs, leading to variability across different datasets. This adaptability stands in contrast to a static approach where the number of attributes is uniformly set across all datasets. To understand the impact of this variability, we explore how altering the number of attributes from 1 to 8 influences model performance. Our findings, detailed in \cref{table:num_experts}, reveal a performance improvement trend as the number of attributes increases, with an optimal peak at 7 attributes before a slight decline at 8. However, crucially, across all fixed-attribute scenarios, none matched the performance achieved through our method's dynamic attribute determination. These results underscore the importance of an adaptive approach to attribute selection, as opposed to a one-size-fits-all strategy.

\noindent \textbf{Design choice of the vision-conditional pooling layer.} Lastly, we ablate the design of the pooling layer, starting from the naive training-free average pooling, to the attention-based pooling mechanism with condition on the input image. Compared to average pooling, VCP demonstrates a performance gain of 1.08\% in the average harmonic mean. Furthermore, when compared with attention-based max pooling, which selects a single description per attribute according to the attention score in \cref{eq:vcp}, VCP maintains a superior advantage of 1.55\% in average harmonic mean. These outcomes attest to the VCP layer's integral role in finetuning attribute relevance to the visual context, substantiating its design and implementation within our model.

\begin{figure}[t]
\begin{minipage}[c]{0.24\textwidth}
\centering
\captionof{table}{Effects of $\alpha$}
\label{table:alpha}
\begin{small}
    \begin{tabular}{c|cc}
        \toprule
        $\alpha$ & 1.0 & 0.4 \\
        \midrule
        Base & 81.54 & \textbf{84.75} \\
        Novel & 73.85 & \textbf{77.63} \\
        HM & 77.51 & \textbf{81.04} \\
        \bottomrule
    \end{tabular}
\end{small}
\end{minipage}\hfill
\begin{minipage}[c]{0.76\textwidth}
\renewcommand\tabcolsep{3pt}
\captionof{table}{Effects of the number of experts.}
\label{table:num_experts}
\centering
\begin{small}
\begin{tabular}{c|cccccccc|c}
\toprule
  $\#$ Attrs. & 1 & 2 & 3 & 4 & 5 & 6 & 7 & 8 & Ours \\
  \midrule
   Base & 83.20 & 83.97& 84.10 & 84.41& 84.45& 84.62& 84.66& 84.74 & \textbf{84.75} \\
   Novel & 74.90 & 76.20 & 76.35& 77.06& 77.13& 77.17& 77.35& 76.67 & \textbf{77.63} \\
   HM & 78.83& 79.90 & 80.04& 80.57& 80.63& 80.72& 80.84& 80.50 & \textbf{81.04}\\
\bottomrule
\end{tabular}
\end{small}
\end{minipage}
\vspace{-15pt}
\end{figure}

\subsection{Visualization}

\textbf{Expert tokens focus on attribute-related regions.}
We further investigate the effects of vision domain experts by visualizing their class activation maps from three illustrative examples using GradCAM~\citep{selvaraju2017grad}, as shown in\cref{fig:vis_interpretablity}. These visualizations underscore the precision with which each expert token concentrates on the image regions pertinent to its designated attribute. Take the first cat image as an example. The ``fur pattern'' expert distinctly highlights the animal's fur texture, whereas the ``ear'' and ``eye'' experts focus precisely on the respective anatomical features.  This pattern of attribute-specific attention is consistent across the evaluated examples, reinforcing the conceptualization of expert tokens as dedicated ``domain experts'' within the visual field.

\begin{figure}[!t]
  \centering
  \includegraphics[width=\textwidth]{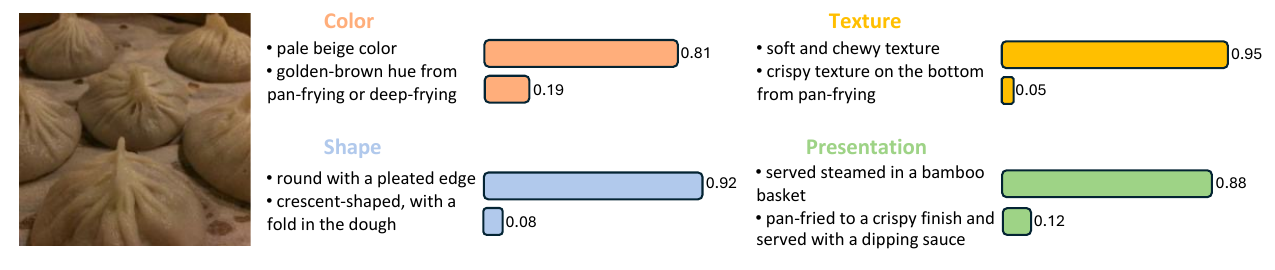}
  \caption{Visualization of the attention weights in the VCP layer for an example ``dumplings'' image.}
   \label{fig:text_interpretablity}
   \vspace{-10pt}
\end{figure}

\begin{table}[t]
\renewcommand\tabcolsep{6pt}
\caption{Ablation on design choice of the VCP layer. Our cross-attention based pooling mechanism demonstrates the best performance among other variants.}
\vspace{-8pt}
\label{table:expert_token}
\begin{center}
\begin{small}
\begin{tabular}{ccc|c}
\toprule
  Pooling Method & Base Acc. & Novel Acc. & HM  \\
  \midrule
   Attn. Max Pooling & 82.90 & 76.36 & 79.49 \\
   Average Pooling & 83.18 & 76.98 & 79.96\\
   \midrule
   VCP (Ours) & \textbf{84.75} & \textbf{77.63} & \textbf{81.04} \\
\bottomrule
\end{tabular}
\end{small}
\end{center}
\vspace{-15pt}
\end{table}

\noindent \textbf{VCP layer pools the most applicable descriptions.} The inherently interpretable nature of the VCP layer, thanks to its attention mechanism, allows for insightful visualizations of its operational process. Through the examination of attention weights assigned by the VCP layer to different attributes in a given image, we elucidate the layer's capability to discern and prioritize the most applicable descriptions. As illustrated in \cref{fig:text_interpretablity} with a ``dumplings'' image, the VCP layer adeptly allocates higher attention weights to descriptions accurately reflecting the observed instance (e.g., assigning weights of 0.92 to ``round with a pleated edge'' under the ``Shape'' attribute and 0.95 to ``soft and chewy texture'' under the Texture''). In contrast, less relevant descriptions for the specific image context (e.g., ``crescent-shaped'' for Shape and ``crispy texture from pan-frying'' for Texture) receive significantly lower weights. This discernment is crucial, given the class dumplings'' encompasses a broad variety of appearances based on cooking methods, yet not all descriptions are fitting for every instance. These visualizations compellingly demonstrate the VCP layer's effectiveness in refining description relevance, thereby enhancing the model's interpretative alignment with the visual content.

\section{Conclusion}
This paper introduces Tree of Attribute Prompt learning (TAP), a novel method that integrates detailed, LLM-generated descriptions within VLMs, achieving state-of-the-art performance in base-to-novel generalization, cross-dataset transfer, and few-shot image classification tasks across 11 diverse datasets. TAP leverages a hierarchical "Tree of Attribute" framework, distilling structured knowledge graphs from LLMs for nuanced representation of visual concepts, and employs learnable "domain expert" tokens and a vision-conditional pooling module for optimal image-text alignment. While promising, we note that the reliance on LLMs presents challenges in fine-grained datasets where similar classes require nuanced differentiation, in which cases LLMs generate identical descriptions for distinct classes, impacting novel class prediction performance. It highlights the current limitations of LLMs in discerning highly fine-grained distinctions. Addressing this challenge through enhanced LLM capabilities or alternative strategies will be a key focus of future research.

\section*{Acknowledgment}

This work was supported in part by Microsoft’s AI for Good Research Lab, the Harvard Data Science Initiative, and NIH Grant R01HD104969.

\bibliography{iclr2025_conference}
\bibliographystyle{iclr2025_conference}

\clearpage
\appendix
\section{Appendix}

\subsection{Model regularization}

Denote the frozen image feature from CLIP vision encoder as $\mathbf{f}^v$, the frozen text feature for description $d$ from CLIP text encoder as $\mathbf{f}^t_d$, and the zero-shot logit prediction from CLIP as ${\hat{y}}$. Additionally, denote the trained image feature as $\mathbf{\Tilde{f}}^v$, the trained text feature for description $d$ as $\mathbf{\Tilde{f}}^t_d$, and the logit prediction from attribute $a$ after training as $\Tilde{y}_a$. The losses are as follows:
\begin{equation}
    L_{L_1-V} = ||\mathbf{f}^v-\mathbf{\Tilde{f}}^v||_1
\end{equation}

\begin{equation}
    L_{con-T}=-\sum_{d\in \mathbb{D}} \bigg (\frac{1}{2} \log\frac{\exp(cos(\mathbf{f}^t_d, \mathbf{\Tilde{f}}^t_d))}{\sum_{k\in \mathbb{D}_s} \exp(cos(\mathbf{f}^t_d, \mathbf{\Tilde{f}}^t_k))} +\frac{1}{2} \log\frac{\exp(cos(\mathbf{f}^t_d, \mathbf{\Tilde{f}}^t_d))}{\sum_{k\in \mathbb{D}_s} \exp(cos(\mathbf{f}^t_k, \mathbf{\Tilde{f}}^t_d))}\bigg )
\end{equation}
\begin{equation}
    L_{{KL-attr}}=\frac{1}{|\mathbb{A}|}\bigg (\sum_{a\in \mathbb{A}} \mathcal{D_{KL}}(\hat{y}, {\Tilde{y}}_a )\bigg )
\end{equation}

\noindent The regularization loss is then:
\begin{equation}
L_{{reg}} = \mu_1 L_{{L_1-V}} + \mu_2 L_{{KL-attr}} + \mu_3 L_{con-T},
\end{equation}

\noindent Our overall training objective is thus given by:
\begin{equation}
    L_{\text{total}}=L_{\text{class}}+L_{\text{reg}}
\end{equation}

To investigate the effectiveness of model regularization, we compare \ours against existing methods with and without regularization. As evidenced in \cref{tab:reg}, the proposed model regularization helps in both base and novel performance, with an increase of $1.62\%$ in average harmonic mean. Comparing to existing methods, \ours is consistently better than other baselines in both settings, demonstrating the robustness of our method.

\begin{table}[h!]
\renewcommand\tabcolsep{6pt}
\caption{Effectiveness of model regularization. \ours achieves favorable results under both settings.}
\label{tab:reg}
\centering
\begin{small}
\begin{tabular}{lccc|c}
\toprule
 & Regularization & Base & Novel & HM \\ 
\midrule
PSRC-reg & $\times$ & 84.21 & 71.79 & 77.51 \\ 
MaPLe & $\times$ & 82.28 & 75.14 & 78.55 \\ 
\rowcolor{tabhighlight}
TAP-reg & $\times$ & \textbf{83.37} & \textbf{75.82} & \textbf{79.42} \\ 
\midrule
PSRC & $\checkmark$ & 84.26 & 76.10 & 79.97 \\ 
\rowcolor{tabhighlight}
TAP & $\checkmark$ & \textbf{84.75} & \textbf{77.63} & \textbf{81.04} \\ 
\bottomrule
\end{tabular}
\end{small}
\end{table}

\subsection{Additional implementation details}

Because the number of attributes vary across the 11 datasets which results in different number of learnable parameters, we group the datasets into two and apply two sets of learning rates to balance generalizability and performance. For DTD, Oxford Flowers, Stanford Cars, UCF101, and Caltech101 datasets, which have fewer attributes, we use a low learning rate of 0.002 for the text encoder to avoid overfitting and a high learning rate of 0.006 for the vision encoder to facilitate the learning process. A high $\mu_3=3$ is also used to regularize the text encoder for preventing overfitting. For the remaining 6 datasets, which have more attributes, the learning rates for both text and vision encoders are set as 0.004, with $\mu_3=1.5$. $\mu_1=10$, and $\mu_2=2.5$ are used for all datasets. 

For base-to-novel generalization and few-shot classification evaluations, we use an adaptive approach for generating the attributes, in which the attributes vary across datasets. Although it turns out to be better than using a fixed set of attributes as shown in \cref{table:num_experts}, it is not applicable to the cross-dataset transfer experiment as both VCP layers and visual expert tokens are specific to their corresponding attributes. Therefore, for cross-dataset transfer, we use the following fixed set of 4 attributes that are applicable to all 11 datasets: Pattern, Texture, Shape, and Context.

We use PyTorch~\cite{paszke2017pytorch} to implement all experiments on a single NVIDIA A100-80GB GPU. Our code is developed based on the implementation of CoOp~\cite{zhou2022learning}, which is available at https://github.com/KaiyangZhou/CoOp and released under the MIT license. Our code is also released under the MIT license. Baseline results for the three tasks are taken from their respective publications. For the ``global context'' attribute which is aligned with the CLS token in the vision encoder, we use the following 7 selected templates provided in~\cite{radford2021learning}. 

\texttt{"itap of a \{class\}."}

\texttt{"a bad photo of the \{class\}."}

\texttt{"a origami \{class\}."}

\texttt{"a photo of the large \{class\}."}

\texttt{"a \{class\} in a video game."}

\texttt{"art of the \{class\}."}

\texttt{"a photo of the small \{class\}."}

\subsection{Robustness of LLMs}

To investigate the robustness of our methods against different LLMs, we additionally generate the descriptions using a locally-served small LLM - Qwen-2-7B-Instruct~\citep{yang2024qwen2}, in which the results are comparable.

\begin{table}[h!]
\renewcommand\tabcolsep{6pt}
\caption{Robustness against different LLMs.}
\label{tab:llm_robustness}
\begin{center}
\begin{small}
\begin{tabular}{ccc|c}
\toprule
  LLMs & Base Acc. & Novel Acc. & HM  \\
  \midrule
   Qwen-2-7B-Instruct & 84.68 & 77.31 & 80.83 \\
   GPT-3.5-Turbo & 84.75 & 77.63 & 81.04 \\
\bottomrule
\end{tabular}
\end{small}
\end{center}
\end{table}

\subsection{Prompts for Tree-of-Attribute generation}
As introduced in Section 3.3, we generate the Tree-of-Attribute with the following three steps: 1) Attribute Generation, 2) In-Context Example Generation, and 3) Description Generation for All Classes. The prompts for each step are as follows:

\textbf{1) Attribute Generation:} 

\textit{\{Dataset Description.\}}

\textit{Visual attributes refer to observable, describable features of the images that can include color, shape, size, texture, and any specific patterns or markings, which can help differentiate between classes for the dataset. They should be consistently observable across multiple images of the same class. Your task is to generate a list of visual attributes (less than 10) for the \{Dataset Name\} dataset. Ensure this list is clear, concise, and specific to the dataset's needs. Avoid generic attributes that do not contribute to distinguishing between classes.}

\textbf{2) In-Context Example Generation}

\textit{Describe describe what a "\{Random Class Name\}" class in the \{Dataset Name\} dataset look like using the generated visual attributes.}

\textit{You must follow the following rules: }

\textit{1. For each visual attribute, describe all possible variations as separate sentences. This approach allows for a detailed and clear presentation of each attribute's range.}

\textit{2. Provide a maximum of five descriptions for each visual attribute to maintain focus and relevance. Also, aim to provide at least two descriptions to ensure a comprehensive overview of the attribute.}

\textit{3. The descriptions should provide clear, distinguishable features of each class to support image classification tasks.}

\textit{4. Descriptions for each attribute are independent from each other, and they should not serve as context for each other.}

\textit{5. Each description describes an image independetly. If certain description is possible for a class, please just list that description, and do not use words like "may have" or "sometimes have".}

\textit{6. Reply descriptions only. Do not include any explanation before and after the description.} 

\textit{7. The descriptions should follow the format of "classname, which ...", where "..." is the description of the visual attribute.}

\textbf{3) Description Generation for All Classes}

\textit{\{Dataset Description.\}}

\textit{Your task is to write detailed descriptions for various classes within the \{Dataset Name\} dataset, using the provided visual attributes such as color and shape. These descriptions will help in accurately classifying and understanding the unique features of each class.}

\textit{You must follow the following rules: }

\textit{1. For each visual attribute, describe all possible variations as separate sentences. This approach allows for a detailed and clear presentation of each attribute's range.}

\textit{2. Provide a maximum of five descriptions for each visual attribute to maintain focus and relevance. Also, aim to provide at least two descriptions to ensure a comprehensive overview of the attribute.}

\textit{3. The descriptions should provide clear, distinguishable features of each class to support image classification tasks.}

\textit{4. Descriptions for each attribute are independent from each other, and they should not serve as context for each other.}

\textit{5. Each description describes an image independetly. If certain description is possible for a class, please just list that description, and do not use words like "may have" or "sometimes have".}

\textit{6. Reply descriptions only. Do not include any explanation before and after the description.} 

\textit{7. The descriptions should follow the format of "classname, which ...", where "..." is the description of the visual attribute.}

\textit{Q: Describe what a "\{Random Class Name\}" in the \{Dataset Name\} look like using the following visual attributes: \{Visual Attributes from Step 1.\}}

\textit{A: \{Answer from Step 2.\}}

\textit{Q: Describe what a "\{Target Class Name\}" in the \{Dataset Name\} look like using the following visual attributes: \{Visual Attributes from Step 1.\}}

\textit{A:}

In the prompt templates, \textit{"Dataset Description"} is the description of the dataset from their official website, \textit{"Random Class Name"} is a randomly sampled class name in the dataset for in-context example generation, and \textit{"Target Class Name"} is the class name of interest for the current query. While step 1 and 2 are made in two consecutive calls to provide contexts which are queried once per dataset, step 3 is queried independently for each of the remaining classes in the dataset. Our carefully designed prompts for step 1 and 2 guide the LLM in generating high-quality examples. Human review further confirms that the generated in-context examples from these prompts are of high quality even without any manual intervention.

\subsection{Attribute sets}

The attribute sets generated by LLMs are shown in \cref{tab:attribute_set_1} - \ref{tab:attribute_set_2}.

\begin{table*}[h!]
    \centering
    \caption{Attribute sets generated by LLMs for the 11 datasets.}
    \begin{tabular}{l|l}
        \toprule
        Dataset & Attributes \\
        \midrule
        \multirow{6}{*}{ImageNet} & Orientation \\
        & Shape \\
        & Pattern \\
        & Texture \\
        & Pose \\
        & Context \\
        \cmidrule{1-2}
        \multirow{5}{*}{Caltech101} & Dominant Feature \\
        & Shape \\
        & Texture \\
        & Color \\
        & Size \\
        \cmidrule{1-2}
        \multirow{6}{*}{StanfordCars} & Body Type \\
        & Wheel Design \\
        & Grille Style \\
        & Headlight Shape \\
        & Rear Taillight Design \\
        & Roof Style \\
        \cmidrule{1-2}
        \multirow{4}{*}{Flowers102} & Color \\
        & Petal \\
        & Center structure \\
        & Stem characteristics \\
        \cmidrule{1-2}
        \multirow{5}{*}{Food101} & Color \\
        & Shape \\
        & Texture \\
        & Ingredients \\
        & Presentation Style \\
    \bottomrule
    \end{tabular}
    \label{tab:attribute_set_1}
\end{table*}

\begin{table*}[h!]
    \centering
    \caption{Attribute sets generated by LLMs for the 11 datasets. Cont.}
    \begin{tabular}{l|l}
        \toprule
        Dataset & Attributes \\
        \midrule
        \multirow{7}{*}{FGVCAircraft} & Wing Configuration \\
        & Winglet Presence \\
        & Engine Configuration \\
        & Number of Engines \\
        & Fuselage Length \\
        & Fuselage shape \\
        & Wingspan \\
        \cmidrule{1-2}
        \multirow{6}{*}{SUN397} & Indoor/Outdoor \\
        & Color \\
        & Dominant elements \\
        & Environment \\
        & Architectural style \\
        & Patterns \\
        \cmidrule{1-2}
        \multirow{4}{*}{DTD} & Texture \\
        & Pattern \\
        & Repetition \\
        & Contrast \\
        \cmidrule{1-2}
        \multirow{7}{*}{EuroSAT} & Contrast \\
        & Texture \\
        & Orientation \\
        & Edge \\
        & Size \\
        & Color \\
        & Symmetry \\
        \cmidrule{1-2}
        \multirow{4}{*}{UCF101} & Action Pose \\
        & Number of People \\
        & Background Setting \\
        & Objects Present \\
    \bottomrule
    \end{tabular}
    \label{tab:attribute_set_2}
\end{table*}

\end{document}